%% file: emo-2021.tex
\documentclass[runningheads]{llncs}

\usepackage{svg}

\newcommand{\orcid}[1]{\href{https://orcid.org/#1}{\includesvg[height = 2ex]{ORCIDiD_iconvector}}}

\usepackage[all]{nowidow}
\usepackage{graphicx}
\usepackage{amsmath}
\usepackage[]{algpseudocode}
\usepackage[colorlinks = true,
            linkcolor = red,
            urlcolor  = cyan,
            citecolor = blue,
            anchorcolor = cyan,bookmarks=false]{hyperref}
\usepackage{subcaption}
\usepackage{float}
\usepackage{url}
\usepackage{multirow}
\usepackage{colortbl}

\usepackage{tikz}
\usepackage[misc]{ifsym}
\begin{document}
\title{Exploring Constraint Handling Techniques in Real-world Problems on MOEA/D with Limited Budget of Evaluations}
\titlerunning{Constraint Handling Techniques in Real-world Problems on MOEA/D}
\author{Felipe Vaz\inst{1}   \and
Yuri Lavinas \inst{2} (\Letter)  \and
Claus Aranha\inst{3} \and
Marcelo Ladeira \inst{4}} 

% /\orcid{0000-0003-2712-5340}
%
\authorrunning{Felipe Vaz et al.}
\institute{
University of Brasilia, Brazil
\email{felipesoueu@live.com} \and
University of Tsukuba, Japan  \email{lavinas.yuri.xp@alumni.tsukuba.ac.jp} \and
	University of Tsukuba, Japan \email{caranha@cs.tsukuba.ac.jp} \and
University of Brasilia, Brazil
\email{mladeira@unb.br}
}
\maketitle              % typeset the header of the contribution

\begin{abstract}
    \input{files/0_abstract}

\end{abstract}

\keywords{Multi Objective Optimization  \and Real World Problems \and Constraint Handling Techniques \and MOEA/D}

\input{files/1_introduction}

\input{files/2_mop}
\input{files/3_chts}
\input{files/4_experiments}
\input{files/5_conclusion}

\bibliographystyle{splncs04}
\bibliography{emo-2021}

\end{document}

%% file: files/0_abstract.tex
% The abstract should briefly summarize the contents of the paper in
% 15--250 words.

Finding good solutions for Multi-objective Optimization (MOPs) Problems is considered a hard problem, especially when considering MOPs with constraints. Thus, most of the works in the context of MOPs do not explore in-depth how different constraints affect the performance of MOP solvers. Here, we focus on exploring the effects of different Constraint Handling Techniques (CHTs) on MOEA/D, a commonly used MOP solver when solving complex real-world MOPs. Moreover, we introduce a simple and effective CHT focusing on the exploration of the decision space, the Three Stage Penalty. We explore each of these CHTs in MOEA/D on two simulated MOPs and six analytic MOPs (eight in total). The results of this work indicate that while the best CHT is problem-dependent, our new proposed Three Stage Penalty achieves competitive results and remarkable performance in terms of hypervolume values in the hard simulated car design MOP.

%and we found out that our new proposed CHT has good hypervolume performance in most of the real-world MOPs.

%% file: files/1_introduction.tex
\section{Introduction}~\label{section:intro}

Multi-objective Optimization Problems (MOPs) appear in many application contexts in which conflicting objective functions need to be simultaneously optimized. Finding good sets of solutions for continuous MOPs is generally considered a hard problem, particularly in real-world MOPs. These real-world MOPs usually have two main characteristics: (1) unknown and irregular shape of the Pareto front and (2) a highly unfeasible (constrained) objective space. These constraints invalidate some solutions, which makes finding a set of feasible solutions a challenging task.

% Real-world MOPs usually have two main characteristics: (1) unknown and irregular shape of the Pareto front and (2) a highly unfeasible (constrained) objective space. These real-world constraints arise from numerous phenomena associated with a problem, such as: physical and functional limitations, and resource availability. These constraints invalidate some solutions, which makes finding a set of feasible solutions a challenging task.

There are several promising multi-objective evolutionary algorithms (MOEAs) that can be used to solve real-world MOPs~\cite{trivedi2017survey,li2009multiobjective,zitzler2004indicator}. They have the ability to modify their behavior when searching for solutions according to the MOP in question taking into account the constraints of each MOP. However, few of these algorithms have their performance evaluated in constrained MOPs or the constraints in the MOPs studied are simple to address~\cite{tanabe2020easy}.

% There are several promising multi-objective evolutionary algorithms (MOEAs) that can be used to solve real-world MOPs~\cite{trivedi2017survey,li2009multiobjective,zitzler2004indicator}. MOEAs are often used to find sets of solutions to optimization problems since they can find solutions to complex problems in a single run. One of the main characteristics of MOEAs is the ability to modify its behavior when searching for solutions according to the optimization problem in question, for example, taking into account the constraints of each MOP. However, few of these algorithms have their performance evaluated in constrained MOPs, and generally, the constraints are simple to address~\cite{tanabe2020easy}.

In this work, we focus on solving real-world MOPs with complicated constraints, using the the Multi-Objective Evolutionary Algorithm Based on Decomposition, MOEA/D~\cite{zhang2007moea}, one of these popular MOEA algorithms. The main features of this algorithm is to decompose the MOP into multiple subproblems and solve them in parallel. Decomposing the MOP in various single-objective subproblems makes the algorithm very flexible for dealing with constraints MOP, because adding a penalty value is straightforward. Given this nature of the single-objective subproblems, MOEA/D can handle well different constraint handling techniques (CHTs) while finding a diverse set of optimal solutions. 

% Recently, several Constraint Handling Techniques (CHTs) have been proposed and studied in the context of single-objective optimization. However, few studies focus on CHTs in multi-objective optimization domain~\cite{datta2014evolutionary,lobato2017multi,tanabe2020easy}. 

This work aims to investigate the use of constraint handling techniques in MOEA/D when solving real-world MOPs and exploring the effects and behavior of different CHTs in real-world problems. For that, we compare real-world analytic MOPs, compiled together by Tanabe et. al~\cite{tanabe2020easy} and two simulated MOPs: (1) the problem of selecting landing sites for a lunar exploration robot~\cite{MoonOrbitingSatellite2015}; and (2) the problem of optimization of car designs~\cite{kohira2018proposal}. To further enhance the performance of MOEA/D, we propose an efficient CHT that works well with problems that require an exploration of the unfeasible search space.

The remainder of this paper is organized as follows: Section~\ref{section:cht}, presents the different types of CHTs that we study in this work as well as our new CHT. Section~\ref{section:comparison} shows the experimental settings, such as which parameters are used in MOEA/D, the evaluation criteria, and the results of the experiments~\ref{section:comparison}. Finally, Section~\ref{section:conclusion} presents our conclusion remarks.

%% file: files/3_chts.tex
\section{Constraint Handling Techniques}~\label{section:cht}

Our goal is to study constraint handling techniques in MOEA/D when solving real-world MOPs yet to be explored in the context of MOEA/D. 

All the constraint handling techniques (CHTs) presented here adapt a constrained MOP into an unconstrained problem, by altering the objectives values according to the CHT under use. They follow the next pattern:

\begin{equation}
       f^{agg}_{penalty}(x)  = f^{agg} + penalty*v(x),
\end{equation}

where $f^{agg}$ is the aggregation function, a principal component of MOEA/D. This aggregation function is used to measure the quality of solutions for each subproblem~\cite{campelo2018moeadr},  generated based on weight vectors using a decomposition method. Since the $f^{agg}$ does not consider the constraint violation, MOEA/D can substitute this aggregation function by the $f^{agg}_{penalty}$, making this EA algorithm able to consider the constraints violation for each solution. The other variables in the equation above is the total violation of the solution, $v(x)$, and one or more penalty factors, named $penalty$.

\paragraph{\textbf{Static CHT.}} This penalty is one of the simplest CHTs. This class of penalty is generally characterized by maintaining its penalty value constant during all the evolutionary process. In this paper, the term ``static penalty'' is used to reference penalties defined by the following equation:

\begin{equation}
       f^{agg}_{penalty}(x)  = f^{agg} + \beta*v(x)
\end{equation}

where $x$ is one solution, $\beta$ is the value of the penalty, and $v$ is the violation.

\paragraph{\textbf{Multi-staged CHT.}} This CHT has different static penalty values for multiple levels of violation (refer to Homaifar et. al~\cite{homaifar1994constrained}). This way, more unfeasible solutions have higher penalties, while almost feasible solutions can be kept in the population for longer, making it possible to achieve a higher exploration. Its main downside is the high number of parameters that depend highly on the problem. Its penalty is defined by:

\begin{equation}
       f^{agg}_{penalty}(x)  = f^{agg} + \sum_{i=1}^{m}(R_{k,i}*max[g_i(x),0]^2)
\end{equation}

Where $R_{k,i}$ are the penalty coefficients, m is the number of constraints, and $k = 1,2,...,l$ where l is the number of levels of violations defined by the user.

\paragraph{\textbf{Dynamic CHT.}} This CHT considers the current generation's number to determine the penalty value. In general, the penalty value is small at the beginning of the search and increases across the generations. The idea is to focus on the diversity of feasible solutions and then later focus on the convergence of those solutions. There are several variations of this penalty, and Joines and Houck \cite{joines1994use} proposed the one used throughout this study. It is defined by:

\begin{equation}
       f^{agg}_{penalty}(x)  = f^{agg} + (C*t)^\alpha*v(x)
\end{equation}

where $C$ and $\alpha$ are constants defined by the user, and $t$ is the number of the current generation.

\paragraph{\textbf{Self-adaptive CHT.}} The main characteristics of this CHT are the absence of any user-defined parameters and the focus on balancing the number of both feasible and unfeasible solutions in the population, favoring unfeasible solutions with good objective functions (for more information, see Tessema and Yen \cite{tessema2007constraint}). This CHT uses the ratio of feasible solutions in the current population to determine the penalty value in a two-penalty based approach. This way, the objective function is modified, having two components: a distance measure and an adaptive penalty. The distance measure changes the aggregated objective function by also considering the effect of the individual's violation:

\begin{equation}
       d(x) =
    \begin{cases}
      v(x), & \text{if}\ r_f=0 \\
      \sqrt{f^{agg}"(x)^2 + v(x)^2}, & \text{otherwise.}
    \end{cases}
\end{equation}

% To calculate $r_f$, the following equation is used:
% \begin{equation}
%     r_f = n_f/n_u,    
% \end{equation}

The rate of feasible solutions n the incumbent population $r_f = n_f/n_u$, where $n_f$ is the number of feasible, $n_u$ is the number of unfeasible and $f^{agg}"$ is the normalized aggregated objective value. The penalty of this CHT is defined by:

\begin{equation}
    p(x) = (1-r_f)M(x) + (r_f)N(x),
\end{equation}

where

\begin{equation}
       M(x)
    \begin{cases}
      0, & \text{if}\ r_f=0, \\
      v(x), & \text{otherwise}
    \end{cases}
\end{equation}

\begin{equation}
       N(x)
    \begin{cases}
      0, & \text{if x is a feasible individual,}\\  
      f^{agg}"(x), & \text{if x is an unfeasible individual}.
    \end{cases}
\end{equation}

\paragraph{\textbf{Three Step CHT.}} In this work, we propose a new penalty-based CHT, the Three Step Penalty. The core idea of this penalty is to get a high exploration of the unfeasible space. This idea comes from the fact that there is a need to search through an unfeasible area to get to better feasible solutions. This penalty has three stages: (1) a non-penalty stage, (2) a low penalty stage, and (3) a high penalty stage. The first stage is responsible for exploring the unfeasible space. The second stage has a low penalty to let the population find the nearby feasible region. At last, the third stage increases the penalty to guarantee the convergence to a feasible region. Its penalty is defined by, where $P_{i}$ is the penalty value of step $i$, and $t$ is the number of the current generation:

\begin{equation}
       f^{agg}_{penalty}(x)  = f^{agg} +
       \begin{cases}
         v(x)*P_{1}, & \text{if $25 > t$} \\  
         v(x)*P_{2}, & \text{if $50 > t \geq 25$}  \\
         v(x)*P_{3}, & \text{if $t \geq 50$} 
    \end{cases}
\end{equation}

% where $P_{i}$ is the penalty value of step $i$, and $t$ is the number of the current generation. %The proportion of the steps are taking into account the total number of generations of 100.

%% file: files/4_experiments.tex
\section{Comparison Study}~\label{section:comparison}

To illustrate the difference in the performance of the different CHT, we compare MOEA/D with the following distinct CHT: (1) Low value and (2) High value Static, (3) Multi-staged, (4) Slow and (5) Fast growth Dynamic, (6) Self-adaptive, and (7) Three Stage Penalty against MOEA/D with no CHT (``No Penalty''). The goal is to verify if there is any improvement, and by how much, in MOEA/D performance when using CHT.

\subsection{Problems}

The real-world analytical multi-objective optimization problems used were selected from the new test suite introduced by Tanabe et. al \cite{tanabe2020easy}: (1) bar truss design (CRE21), the objectives are to minimize the structural weight and displacement resulting from the joint; (2) design of welded beams (CRE22), the objectives are to minimize the cost and the final deflection of the welded beams; (3) disc brake design (CRE23), the objectives are to minimize brake mass and decrease downtime; (4) side impact design of the car (CRE31), the objectives are to minimize the average weight of the car, the average speed of the column \textit{V} responsible for supporting the impact load and the force experienced by a passenger; (5) conceptual submarine project (CRE32), the objectives are to minimize the cost of transportation, the weight of the light vessel and the annual cargo transport capacity; (6) water resource planning (CRE51), the objectives are to minimize the cost of the drainage network, the cost of installing the storage center, the cost of installing the treatment center, the expected cost of flood damage and the expected economic loss due to the flood.

%CRE21, CRE22, CRE23, CRE31, CRE32, CRE51. The real-world simulated MOPs used are the three recently proposed by the Japanese evolutionary computing society: (1) Mazda benchmark problem (MAZDA)~\cite{kohira2018proposal} and the (2) lunar landing site selection (MOON)~\cite{MoonOrbitingSatellite2015}.

 The real-world simulated MOPs used are the three recently proposed by the Japanese evolutionary computing society: (1) Mazda benchmark problem (MAZDA): this is a discrete optimizing problem with the goal of designing MAZDA cars, where the objectives are to maximize the number of parts common to three different cars and minimize the total weight of the car~\cite{kohira2018proposal}; (2) lunar landing site selection (MOON): the goal is to select landing sites coordinates (x,y) for a lunar exploration robot to land with the objectives of minimizing the number of continuous shaded days, minimizing the inverse of the total communication time~\footnote{i.e. maximizing the total communication time.}, and minimizing the tilt angle~\cite{MoonOrbitingSatellite2015}.

\subsection{Experimental Parameters}~\label{parameters}

\input{files/parameter_Table.tex}

We use the chosen components of MOEA/D as introduced by Li and Zhang~\cite{zhang2007moea}. The parameters of the crossover and mutation operators and the probability of using the neighborhood with these operators were fine-tuned using irace~\cite{bezerra2016automatic,lopez2016irace} using the the DTLZ unconstrained problems~\cite{deb2005scalable}. We make these parameter choices to isolate the contribution of each CHT. Table~\ref{chap4:parameter_table} summarizes the experimental parameters.

Details of these parameters can be found in the documentation of package MOEADr and the original MOEA/D reference~\cite{zhang2007moea,moeadr_package,moeadr_paper}. All objectives and violations were linearly scaled at every iteration to the interval $\left[0,1\right]$. The scaling of the violations uses the maximum and minimum constraint violations of the new and the incumbent populations as reference for the scaling. Finally, the Weighted Tchebycheff scalarization function \cite{miettinen2008introduction} was used.

% The number of evaluations for all problems was selected based on the recommended number of evaluations on the simulated problem competitions. Because of that, the number of evaluations used throughout this is study is set at 30000.

\input{files/cht_Table.tex}

The parameters of the CHTs can be found in Table~\ref{chap4:cht_table}. There are two instances of dynamic penalty to test the effects of slow growth and high growth of the penalty value. We name these two instances as Slow growth and Fast growth Dynamic Penalty, respectively. There are also two static penalty cases. The first one, called Low value Static, has a low penalty value ($\beta=1$), aiming for a more balanced search and the second one, called High value Static, has a high penalty value ($\beta=1000$), that greatly penalizes unfeasible solutions. We recall the reader that since the self-adaptive penalty requires no parameters, this CHT is not included in Table~\ref{chap4:cht_table}.

\subsection{Experimental Evaluation}~\label{evaluation}

We compare the different strategies using the Hypervolume (HV, higher is better) indicator. For reproducibility purposes, all the code and experimental scripts are available online~\footnote{\label{gitnote}\href{https://github.com/LordeFelipe/MultiObjectiveProjects}{https://github.com/LordeFelipe/MultiObjectiveProjects}}.

For the calculation of HV for the real analytical problems, the objective function was scaled to the $(0,1)$ interval, with reference points set to $(1.1,1.1)$, for two-objective problems; and $(1.1,1.1,1.1)$, for three-objective ones. For the real-world simulated MOPs, the reference point was the recommended in the competitions of the Japanese Society for Evolutionary Computation $(1.1,0)$ for the two-objective car design problem~\cite{Competition2017} and $(1,0,1)$ for the three-objective moon landing problem~\cite{Competition2018}.

\subsection{Experimental Results}~\label{hv_results}

\input{files/creHV_Table.tex}

\input{files/mazdamoonHV_Table.tex}

Here, we present and discuss how is the performance in the real-world constrained problems described above of the following CHT: Low-value Static, High value Static, Slow growth Dynamic, Fast growth Dynamic, Multi-staged, Self-adaptive and Three Stage penalties, and for control, we evaluate the performance of MOEA/D with no CHT (``No penalty'').

\subsection{Results of the CHTs on the Analytical Problems}

Table~\ref{chap4:cre_hv_table} shows the mean hypervolume (HV, higher is better) of the final population of each CHT on the analytical Benchmark Problems. The results show that the performance of MOEA/D can be improve by any CHT, with no clear difference in HV values between CHTs in this experiments. Because of that, it seems that any focus on the exploration of the feasible space is sufficient for drastically improving the performance of MOEA/D in these problems.

\subsection{Results of the CHTs on the Simulated Problems}

Now, we describe our results on the MAZDA problem and MOON benchmark problems in Table~\ref{chap4:mazda_hv_table}. In contrary to the results of the CHTs on the analytical problems, here, it is noticeable that the the right choice of the CHTs has a deep impact on the performance of MOEA/D. This fact suggests that these real-world simulated problems are harder to solve, when compared to the analytical benchmark problems and that finding the right technique to explore the feasible search space has high relevance. Given the performance of MOEA/D in terms of hypervolume, the best CHT for the MAZDA benchmark problem is the Three-Stage Penalty, our newly proposed technique, and the best for the moon landing problem is the Fast growth Dynamic Penalty.

% Both Figure ~\ref{fig:moonHV} and~\ref{fig:mazdaHV} make it clear that using a high static penalty limits the growth of the HV, contrary to the results from the analytical problems. Although the HV is slow to grow initially due to the lack of feasible solutions, it increases very quickly compared to non-exploratory techniques.  

\subsection{Hypervolume Anytime Performance of MOEA/D with CHT}\label{chap4:anytime}

\begin{figure*}[htbp]
  \includegraphics[width=1\textwidth]{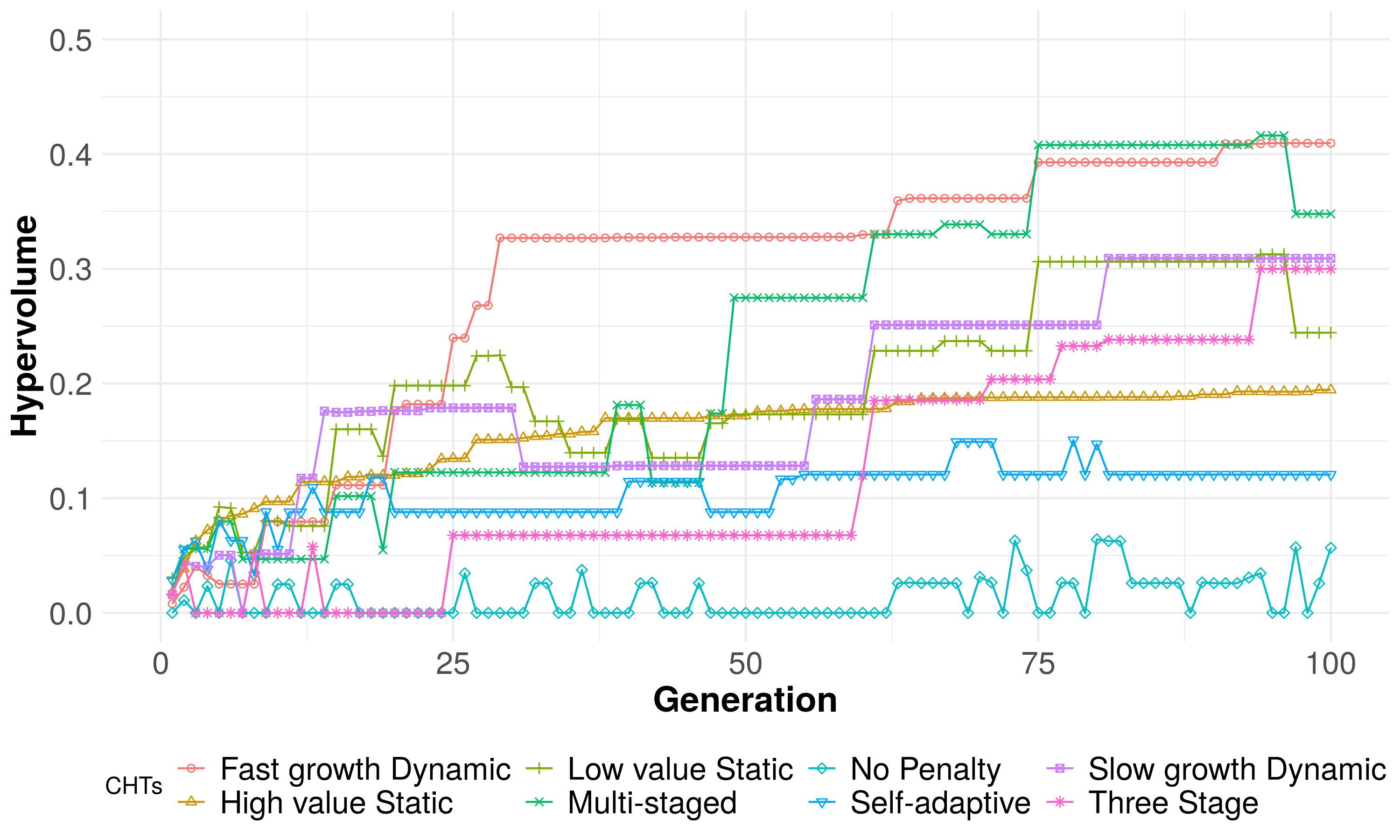}
  \caption{Anytime HV (higher is better) performance of the different CHT in MOEA/D in the moon landing problem. The Three Stage CHT has the competitive performance with most of the other CHT, but is overcomed by the Fast growth Dynamic CHT.}
  \label{fig:moonHV}
\end{figure*}

\begin{figure*}[htbp]
  \includegraphics[width=1\textwidth]{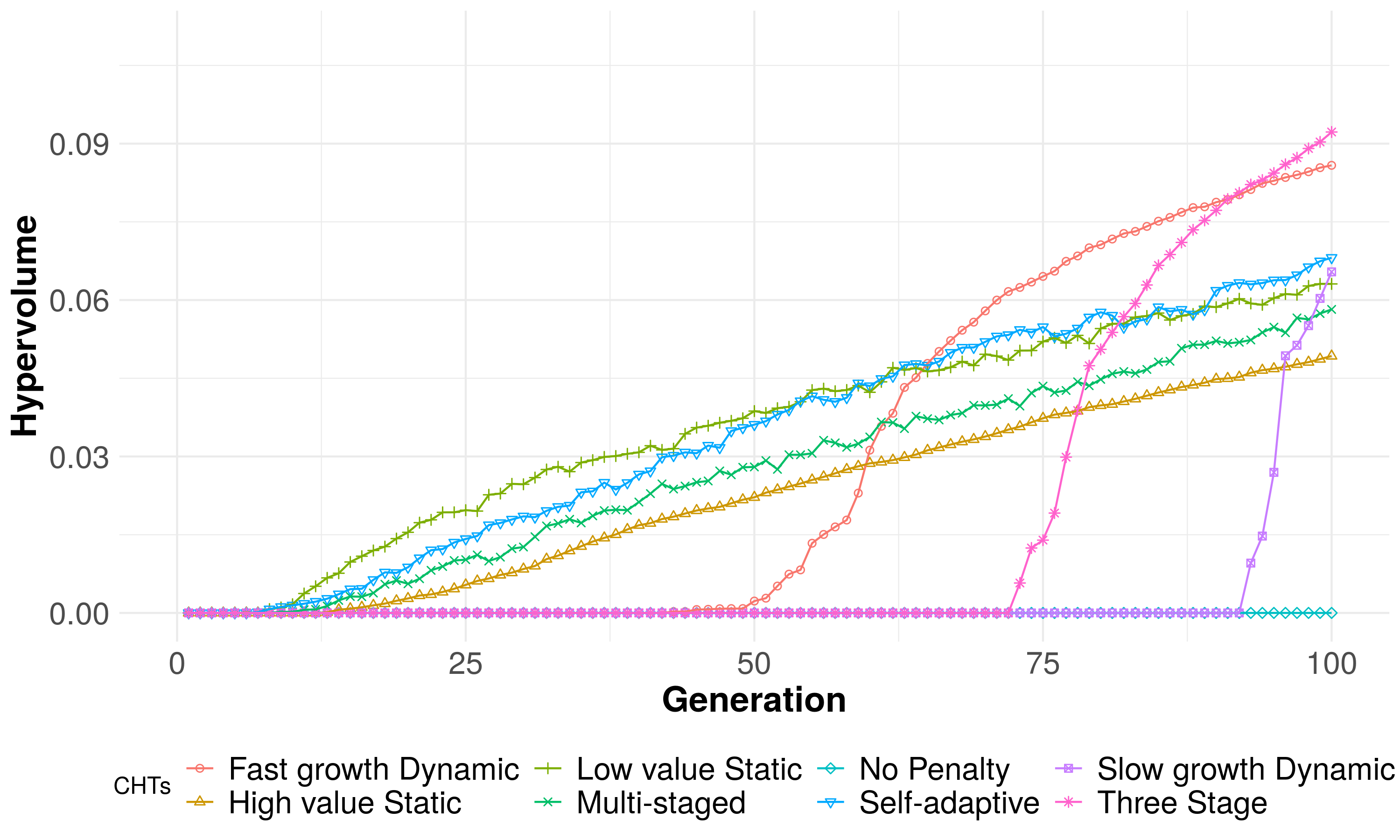}
  \caption{Anytime hypervolume (HV, higher is better) performance of the different CHT in MOEA/D in the MAZDA car problem. The Three Stage CHT achieves the best values at the end of the search execution.}
  \label{fig:mazdaHV}
\end{figure*}

Here we show the anytime performance of the different CHT in terms of HV values. Figures~\ref{fig:moonHV} and~\ref{fig:mazdaHV} illustrate the hypervolume anytime (higher is better) of the eight different MOEA/D variants (seven with different CHT and one without CHT) for the MOON and MAZDA problems, respectively. These results indicate that, at the start of the search (in the first few generations), the static, self-adapting, and multi-staged CHTs have a better performance. Our interpretation for such results comes from the observation that these CHTs put more pressure on the population to get feasible solutions right from the beginning. 

On the other side, dynamic and three-stage penalties initially have a poor performance but get better quickly when the penalty value rises, thus increasing the pressure towards a population with more feasible solutions. We understand that these Figures (\ref{fig:moonHV} and~\ref{fig:mazdaHV}) suggest that increasing the pressure at the right generation may have a deep impact on MOEA/D's hypervolume performance in the simulated problems studied here. 

As we can see in Figure~\ref{fig:mazdaHV}, all algorithms failed to converge adequately, and their performance might still improve with more generations. This might also be true for the MOON problem, since Figure~\ref{fig:moonHV} does not show clear indicatives that all variants of MOEA/D with the different CHT have converged, since at least some of them are still having small changes in HV values close to the final generations.
 

%% file: files/parameter_table.tex
\setlength{\tabcolsep}{8pt}
\begin{table}[htb]
\centering
	\normalsize
	\caption{Experimental parameter settings.}
	\label{chap4:parameter_table}
    \begin{tabular}{|l|l|}
        \hline
        \rowcolor[gray]{.85}MOEA/D parameters             & Value            \\ \hline
        \multirow{6}{*}{Polynomial mutation params.}        & $\eta_m = 56$ (2 obj)    \\
                                                            & $\eta_m = 26$ (3 obj)    \\
                                                            & $\eta_m = 31$ (5 obj)    \\ &  --------------------------------------------\\ 
                                                            & $p_m = 0.55$ (2 obj)    \\
                                                            & $p_m = 0.15$ (3 obj)    \\
                                                            & $p_m = 0.05$ (5 obj)    \\ \hline
        \multirow{6}{*}{Simulated Binary Crossover (SBX).}  & $\eta_m = 24$ (2 obj)    \\
                                                            & $\eta_m = 74$ (3 obj)    \\
                                                            & $\eta_m = 29$ (5 obj)    \\ &  --------------------------------------------\\ 
                                                            & $p_c = 0.7$ (2 obj)    \\
                                                            & $p_c = 0.65$ (3 obj)    \\
                                                            & $p_c = 0.7$ (5 obj)    \\ \hline
        Neighborhood size                                   & $T = 20\%$ of the population size        \\
        \hline
        \multirow{3}{*}{Neighborhood probability}     & $\delta = 0.981$ (2 obj) \\
                                                            & $\delta = 0.471$ (3 obj) \\
                                                            & $\delta = 0.971$ (5 obj) \\ \hline
        \multirow{3}{*}{SLD decomposition param.}           & $h = 299$ (2 obj)  \\
                                                            & $h = 14$ (3 obj) \\
                                                            & $h = 8$ (5 obj) \\ \hline
        \multirow{2}{*}{Population size}                    & $N = 300$ (2 and 3 obj) \\
                                                            & $N = 210$ (5 obj) \\ \hline

        \multicolumn{2}{c}{}\\
        \hline
        \rowcolor[gray]{.85}Experiment Parameters           & Value            \\ \hline
        Repeated runs                                       & 21               \\ \hline
        Computational budget                                & 30000 evaluations \\ \hline
\end{tabular}
\end{table}

%% file: files/cht_table.tex
\setlength{\tabcolsep}{8pt}
\begin{table}[htb]
\centering
	\normalsize
	\caption{Experimental CHT parameters.}
	\label{chap4:cht_table}
    \begin{tabular}{|l|l|}
        \hline
        \rowcolor[gray]{.85}CHT parameters                  & Value            \\ \hline
        Low value Static                                      & $\beta = 1$     \\ \hline
        High value Static                                      & $\beta = 1000$     \\ \hline
        \multirow{4}{*}{Dynamic}                    & $C = 0.05$     \\ 
                                                            & $\alpha = 2$     \\ & 
                                                            --------------------------------------------\\ 
                                                            & $C = 0.02$     \\ 
                                                            & $\alpha = 2$     \\  \hline
        \multirow{4}{*}{Multi-staged}                & $R_1 = 1$ if $v(x) < 0.25$     \\
                                                            & $R_2 = 2$ if $0.5 > v(x) \geq 0.25$    \\ 
                                                            & $R_3 = 4$ if $0.75 > v(x) \geq 0.5$    \\ 
                                                            & $R_4 = 8$ if $v(x) \geq 0.75$    \\ \hline
        \multirow{3}{*}{Three Step}               & $P_1 = 0$ if $25 > t$     \\
                                                            & $P_2 = 1$ if $50 > t \geq 25$    \\ 
                                                            & $P_3 = 1000$ if $t \geq 50$    \\ \hline
                                                            
\end{tabular}
\end{table}

%% file: files/creHV_table.tex
\setlength{\tabcolsep}{4pt}
\begin{table}[htb]
\centering
	\normalsize
	\caption{Hypervolume mean results of the analytical problems. Best HV values are highlighted in bold. It is clear that for this set of problems, using any CHT is the best choice over MOEA/D with no CHT penalty (``No penalty''). The newly proposed Three Stage CHT performs the best in the CR21, CR22 and CRE31 with competitive performance on all of the other problems.}
	\label{chap4:cre_hv_table}
    \begin{tabular}{|l|l|l|l|l|l|l|}
        \hline
        \rowcolor[gray]{.85}CHT & CRE21 & CRE22 & CRE23 & CRE31 & CRE32 & CRE51   \\ \hline
        No penalty                              & 0.00 & 0.05 & 0.84 & 0.75 & 0.00 & 0.88 \\ \hline
        Low value Static                          & 1.09 & \textbf{1.20} & 1.16 & \textbf{0.81} & 0.99 & \textbf{1.01} \\ \hline
        High value Static       & 1.09 & \textbf{1.20} &  \textbf{1.18} & \textbf{0.81} & \textbf{1.01} & 1.00 \\ \hline
        Multi-staged                             &  1.09 & \textbf{1.20} & 1.17 & \textbf{0.81} & 0.99 & \textbf{1.01} \\ \hline
        Slow growth Dynamic           & 1.09 & \textbf{1.20} & 0.98 & \textbf{0.81} & \textbf{1.01} & 1.00 \\ \hline
        Fast growth Dynamic           & 1.07 & \textbf{1.20} & 1.12 & \textbf{0.81} & 1.00 & 1.00 \\ \hline
        Self-adaptive                            & 1.10 & \textbf{1.20} & 1.16 & \textbf{0.81} & 1.00 & 1.00 \\ \hline
        Three Stage                     & \textbf{1.11} & \textbf{1.20} & 1.11 & \textbf{0.81} & 0.82 & 0.99 \\ \hline

\end{tabular}
\end{table}

%CRE31
%3stageNEW  dynamicC002alpha2   dynamicC005alpha2   multistaged     none        selfadapting    static1     static1000
%0.8084083  0.8127528           0.8128627           0.8158512       0.7530823   0.8139333       0.8154623   0.8158015

%CRE32
%3stageNEW  dynamicC002alpha2   dynamicC005alpha2   multistaged     none        selfadapting    static1     static1000
%0.8240670  1.0127370           0.9961298           0.9945692       0.0000000   0.9999663       0.9945692   1.0092386

%CRE51
%3stageNEW  dynamicC002alpha2   dynamicC005alpha2   multistaged     none        selfadapting    static1     static1000
%0.9972770  1.0066407           1.0064282           1.0110701       0.8801649   1.0083562       1.0110701   1.0092114

%CRE21
%3stageNEW  dynamicC002alpha2   dynamicC005alpha2   multistaged     none        selfadapting    static1     static1000
%1.111998   1.094070            1.075955            1.093961        0.000000    1.102019        1.093961    1.090267

%CRE22
%3stageNEW  dynamicC002alpha2   dynamicC005alpha2   multistaged     none        selfadapting    static1     static1000
%1.20393319 1.19973739          1.19995646          1.19995693      0.04930816   1.19918069     1.19963016  1.20546557

%CRE23
%3stageNEW  dynamicC002alpha2   dynamicC005alpha2   multistaged     none        selfadapting    static1     static1000
%1.1104124  0.9853830           1.1175967           1.1714406       0.8416941   1.1654387       1.1575480   1.1767306

%% file: files/mazdamoonHV_table.tex
\setlength{\tabcolsep}{12pt}
\begin{table}[htb]
\centering
	\normalsize
	\caption{Hypervolume mean results of the simulated problems. Best HV values are highlighted in bold and in case of ties, the standard deviation was used. For the MAZDA problem this best CHT is the Three State Penalty followed closely by the Fast growth Dynamic Penalty. For the MOON problem, the best CHT is the Fast growth Dyamic Penalty.}
	\label{chap4:mazda_hv_table}
    \begin{tabular}{|c|c|c|}
        \hline
        \rowcolor[gray]{.85}CHT                 & MAZDA  & MOON   \\ \hline
        No penalty                              & $0.00 \pm 0.00 $ & $0.06 \pm  0.18 $     \\ \hline
        Low value Static                           & $0.06 \pm 0.01 $ & $0.24 \pm 0.32 $     \\ \hline
        High value Static       & $0.05 \pm 0.01 $ & $0.19 \pm 0.10 $     \\ \hline
        Multi-staged                             & $0.06 \pm 0.02 $ & $0.35 \pm 0.34 $     \\ \hline
        Slow growth Dynamic           & $0.06 \pm 0.02 $ & $0.31 \pm 0.33 $     \\ \hline
        Fast growth Dynamic            & $0.09 \pm 0.03 $ & \pmb{$0.41 \pm 0.31 $}     \\ \hline
        Self-adaptive                            & $0.07 \pm 0.02 $ & $0.12 \pm 0.23 $     \\ \hline
        Three Stage                      & \pmb{$0.09 \pm 0.01 $} & $0.30 \pm 0.31 $     \\ \hline

\end{tabular}
\end{table}

%3stageNEW  dynamicC002alpha2   dynamicC005alpha2   multistaged     none        selfadapting    static1     static1000
%0.09223019 0.06540491          0.08583730          0.05820700      0.00000000  0.06811722      0.06309363  0.04926876
%0.01299714 0.02398967          0.02730981          0.01718936      0.00000000  0.01585445      0.01330083  0.01471058

%% file: files/5_conclusion.tex
\section{Conclusion}~\label{section:conclusion}

In this work, we study seven different CHT in both analytical and simulated real-world problems. It is noticeable that using CHTs can be responsible for increments in hypervolume values compared to traditional MOEA/D without penalty. It comes with no surprise that while a CHT may be a reasonable choice for a given multi-objective problem (MOPs), this CHT might still perform poorly in other MOPs, with different characteristics. Moreover, most CHT are parameter sensitive, which is also affected by the MOP of choice and parameter choice. That said, the results of this work suggest that both the Three Stage and the Dynamic CHTs are a good choice for finding optimal feasible solutions and for improving the performance of MOEA/D in most constrained real-world MOP study in this work. We highlight the increase of performance showed by our newly proposed penalty, particularly in the real-world simulated car design.

The Three Stage CHT has few parameters and we recall the reader that the idea behind this technique is simple yet efficient: first, the Three State CHT lets MOEA/D have a free exploration the search space, with no penalty to solutions that violate constraints, them this CHT increases the penalty of solutions that violate constraints to an intermediate value, forcing MOEA/D to prefer solutions that are feasible while keeping good performing unfeasible solutions to only at the final stage start to heavily penalize unfeasible solutions and given high preference to feasible solutions. This way MOEA/D can further explore the search space while being able to select optimal feasible solutions at the end of the run. 

Future works include finding ways to improve the performance of MOEA/D on simulated real-world problems using MOEA/D variants that use resource allocation (RA)  \cite{lavinas2019using,lavinas2019improving,lavinas2020moea} since they show promising results in MOP without constraints. We also plan to study applying CHT as the method for RA as it can control the allocation of resources to explore the unfeasible space more efficiently.